\documentclass[runningheads]{llncs}
\usepackage[main=english,thai]{babel}
\usepackage[utf8x]{inputenc}
\usepackage{enumitem}
\usepackage{listings}
\usepackage{threeparttable}
\usepackage{multirow}
\usepackage{graphicx}
\usepackage[caption=false]{subfig}
\usepackage{etoolbox,siunitx}
\robustify\bfseries
\usepackage{booktabs, makecell}
\begin{document}
\title{A Comparative Study of Pretrained Language Models on Thai Social Text Categorization}
\titlerunning{Comparative Study of Pretrained Language Models}
\author{Thanapapas Horsuwan\inst{1} \and Kasidis Kanwatchara\inst{1} \and \\ Peerapon Vateekul\inst{1} \and Boonserm Kijsirikul\inst{1}}

\authorrunning{T. Horsuwan et al.}
\institute{Department of Computer Engineering, Faculty of Engineering, \\ Chulalongkorn University, Bangkok, Thailand\\
\email{\{thanapapas.h,kanwatchara.k\}@gmail.com}\\
\email{\{peerapon.v,boonserm.k\}@chula.ac.th}\\
}
\maketitle              % typeset the header of the contribution
\begin{abstract}

The ever-growing volume of data of user-generated content on social media provides a nearly unlimited corpus of unlabeled data even in languages where resources are scarce. In this paper, we demonstrate that state-of-the-art results on two Thai social text categorization tasks can be realized by pretraining a language model on a large noisy Thai social media corpus of over 1.26 billion tokens and later fine-tuned on the downstream classification tasks. Due to the linguistically noisy and domain-specific nature of the content, our unique data preprocessing steps designed for Thai social media were utilized to ease the training comprehension of the model. We compared four modern language models: ULMFiT, ELMo with biLSTM, OpenAI GPT, and BERT. We systematically compared the models across different dimensions including speed of pretraining and fine-tuning, perplexity, downstream classification benchmarks, and performance in limited pretraining data.

%With a generalized language model that characterizes colloquial language on social media, we are able to obtain new state-of-the-art results on Wongnai Rating Review Predictions Challenge with test micro-F1 of 62.51 (3.20 absolute improvement). 

\keywords{language model \and pretraining  \and Thai social media \and comparative study \and data preprocessing}
\end{abstract}

%--- Begin Sections---
\section{Introduction}
% PURPOSE: Why should we do this work?
Social networks are active platforms rich with a quickly accessible climate of opinion and community sentiment regarding various trending topics. The growth of the online lifestyle is observed by the bustling active communication on social media platforms. Opinion-oriented information gathering systems aim to extract insights on different topics, which have numerous applications from businesses to social sciences. Nevertheless, existing NLP researches on utilizing these abounding noisy user-generated content have been limited despite its potential value.

First introduced in \cite{dai2015}, pretrained language models (LMs) have been a topic of interest in the NLP community. This interest has been coupled with works reporting state-of-the-art results on a diverse set of tasks in NLP. In light of the notable benefits of transfer learning, we chose to compare four renowned LMs: ULMFiT \cite{ulmfit}, ELMo with biLSTM \cite{elmo}, OpenAI GPT \cite{openai-gpt}, and BERT \cite{bert}. To the best of our knowledge, our work is the first comparative study conducted on pretrained LMs in Thai language. Our LMs were trained in a three-stage process as per suggested in \cite{ulmfit}: LM pretraining, LM fine-tuning, and classifier fine-tuning. The goal of unsupervised pretraining is to find a good initialization point to capture the various general meaningful aspects of a language. Befitting Thai language with resource scarcity, we expect that pretraining user-generated content would serve as a solid basis for transfer learning to downstream tasks. 

% < Talk about why are we going to use the Pantip.com >
Pantip is the largest Thai internet forum with a huge active community where a diverse range of topics are discussed. The variability of surplus examples from Pantip covers the basic linguistic syntax of Thai language while maintaining the colloquial and noisy nature of online user-generated content. In this paper, we investigate and compare the capability of each LM to capture the relevant features of a domain-specific language via pretraining copious unlabeled data from user-generated content.

% LIST CONTRIBUTIONS
The main contributions of this paper are the following: 
\begin{itemize}
  \item We developed unique data preprocessing techniques  for Thai social media. 
  \item We pretrained ULMFiT, ELMo, GPT, and BERT on a noisy Thai social media corpus much larger than the existing Thai Wikipedia Dump. 
  \item We compared the language models across different dimensions including speed of pretraining and fine-tuning, perplexity, downstream classification benchmarks, and performance in limited pretraining data.
  \item Our pretrained models and code can be obtained upon request to the corresponding authors
\end{itemize}

This paper is organized as follows. Our data preprocessing techniques are explained in Section~\ref{section:data_preprocessing} and the LMs used for pretraining are briefly described in Section~\ref{section:pretrain_lm}. The datasets used in this paper are described in Section~\ref{section:dataset} and Section~\ref{section:setup} explains our hyperparameters and evaluation metrics. The results are reported in Section~\ref{section:result} and finally concluded in Section~\ref{section:conclusion}.

\section{Our Data Preprocessing for Thai Social Media} \label{section:data_preprocessing}
Data preprocessing is one of the most important phases in improving the learning comprehension of the LMs. If much irrelevant and redundant information introduces unwanted noise in the training corpus, it is difficult for the models to discover knowledge during the training phase. This is especially true for unfiltered data from user-generated content on social media, where it requires specific methods of data preprocessing unique to the domain.

% NEED TO TALK ABOUT HOW NOISY IS THE DATA %% and, later talk about how we are going to mitigate it
The Thai webboard Pantip allows members to freely create threads as long as it conforms to a list of actively regulated etiquette.
% Members who are interested in that topic would join the discussion in the comments section. 
The colloquial nature of the data posted allows for huge amounts of noise to be introduced in the data, such as ASCII arts, language corruption 
\foreignlanguage{thai}{(ภาษาวิบัติ),}
 irregular spacing, misspelling, character repetition, and spams \cite{ekapolUGWC}. The unpredictable noise in the data substantially increases the vagueness of word boundary, which already is a problem in formal Thai language \cite{wordseg-thai}. Additionally, Thai word segmentation is dependent on context. A famous example is the compound word \foreignlanguage{thai}{`ตากลม'}, which can be either split into \foreignlanguage{thai}{`[ตา][กลม]'} or \foreignlanguage{thai}{`[ตาก][ลม]'}. Both are grammatically correct when used within their corresponding context. To ease the impact of the issues, the data preprocessing approaches we employed are as follows:
\begin{enumerate}
    \item \textbf{Length Filtering} To select meaningful threads to the LM, threads with a title and with a body of more than 100 characters were selected. 
    ~
    \item \textbf{Language Filtering} An n-gram-based text categorization library langdetect\footnote{github.com/fedelopez77/langdetect} was used to filter out the threads that are not labeled as Thai language. 
    ~
    \item \textbf{General Preprocessing before Tokenization} Inspired by \cite{ulmfit,thai2fit}, the techniques include fixing HTML tags, removing duplicate spaces and newlines, removing empty brackets, and adding spaces around `/' and '\#'. In addition, character order in Thai language may be typed in a different sequence but visually rendered in the same way. This is due to the fact that vowels, diphthongs, tonal marks, and diacritics may sometimes be physically located above or below the base glyph--allowing different sequential orders to appear visually equivalent. Thus, normalizing the character order is required for the machine to understand the seemingly similar tokens.
    ~
    \item \textbf{Customized Preprocessing before Tokenization} We also developed and customized techniques suitable for Thai social media. 
    Last character repetition is a common behavior of Thai people analogous to prolonging the vowel sounds of a word in spoken language to emphasize certain emotions. We truncate the word and follow it by a special token. pyThaiNLP \cite{pythainlp} adopted a similar technique but we implemented minor modifications of space addition following the token for better tokenization results. 
    %Special tokens are utilized for character repetitions due to the Thai common behavior of last character repetition. This is analogous to prolonging the vowel sounds of a word in spoken language to emphasize certain emotions. pyThaiNLP \cite{pythainlp} adopted a similar technique but we implemented minor modifications to add a space following the token for better tokenization results. 
    Likewise, a special token is used for word repetitions similar to \cite{ulmfit} preprocessing technique, which at the time this technique has not been widely used in Thai language preprocessing. Since Thai is a language without word boundaries, our algorithm recognizes words as any character sequence of more than 2 characters with more than 2 repetitions of that sequence. 
    All types of repetitions are truncated to 5 as it provides no higher emotional impact and to limit the vocabulary size. 

    In addition, we also propose 2 new preprocessing methods: a special token for any numeric strings and a special token for laughing expressions.
    
    We replaced all strings related to numbers with a special token: general numbers, masked and unmasked phone numbers, Thai numbers, date and time, masked prices, and numbers of special forms. Although differentiating the numbers provide some semantic value, the sparsity of the information would most likely make these numbers tail out of vocabulary (OOV) tokens. We believe that this preprocessing method would allow the language models to more generally understand how numbers are used in text.

    In an online environment, Thai people often express laughter in written language with an onomatopoeia, utilizing the repetition of `5' followed by an optional `+'. This is due to the fact that the Thai pronunciation of `5' is `ha'. We replaced all tokens with more than 3 consecutive `5' and an optional `+' with a special laugh token. Although this may have a minor effect on actual numbers, this onomatopoeia is very commonly used in Thai online context and it is important for the model to learn this special token. An example is provided in Table~\ref{table:data_pre_ex} for clarification.

\begin{table}
\centering
\sisetup{round-mode=places,detect-weight=true,detect-inline-weight=math}
\caption{An example of our preprocessing method. \texttt{[CREP]} and \texttt{[LAUGH]} are special tokens used for character repetition and laughing respectively.} \label{table:data_pre_ex} 
    \begin{threeparttable}
        \begin{tabular}{ c | c c }
        \toprule
        & Before & After  \\
        \midrule
        Thai &\foreignlanguage{thai}{ฉันชอบมันมากกกก555555+}   & \foreignlanguage{thai}{ฉันชอบมันมาก} 
        \texttt{[CREP]} 4 \texttt{[LAUGH]}\\
        Translated & I like it a lotttt hahahahaha\tnote{1}& I like it a lot \texttt{[CREP]} 4 \texttt{[LAUGH]}\\
        \bottomrule
        \end{tabular}
    \begin{tablenotes}
    \item[1] { `5' is pronounced `ha' in Thai}
    \end{tablenotes}
    \end{threeparttable}
    \vspace{-6mm}%Put here to reduce too much white space after your table 
\end{table}
    
    ~
    %\cite{pythainlp, thainer, onlineenglish, khwamruphasathai, aegisub-thai-dict, wordcutpy}
    \item \textbf{Tokenization} We used the pyThaiNLP \cite{pythainlp} default tokenizer, which is a dictionary-based Maximum Matching with Thai Character Cluster. However, we created our own aggregated dictionary for tokenization to improve the tokenization accuracy for colloquial user-generated content. The dictionary\footnote{The dictionary is referenced at our GitHub \url{https://github.com/Knight-H/thai-lm}} is compiled from various sources of data, including general words, abbreviations, transliterations, named entities, and self-annotated Thai slangs and commonly used corrupted language. This includes word variants like 
    \foreignlanguage{thai}{ฮะ ฮาฟ ฮับ ฮัฟ คร้าบ ค้าฟ ค้าบ คับ ครัช}
    which are all word variants of the suffix to indicate formality
    \foreignlanguage{thai}{ครับ.}
     The vocabulary is built from the most common 80k tokens. 
    ~
    \item \textbf{General Preprocessing after Tokenization:} Following \cite{ulmfit} and \cite{thai2fit}, some general preprocessing techniques after tokenization were used. This includes ungrouping the emoji's from text, and to lowercase all English words. 
    ~
    %\cite{nongbot, wiki-wrongs, slang1, slang2, itwords}
    \item \textbf{Spelling Correction} In an effort to reduce the number of unnecessary tokens sprouting from incorrectly spelled words, we compiled a list of commonly misspelled word mappings aggregated from various sources. We corrected and standardized the vocabulary used. This is an important task due to the free and lax nature of the corpus, where a single word may be represented in different variants or misspelled and abbreviated into various tokens. Note that not all replacements can be made due to the collision of actual vocabularies and the limited comprehensiveness of the list. 
\end{enumerate}

\section{Pretrained Language Models in Our Study} \label{section:pretrain_lm}
\subsection{Universal Language Model Fine-tuning (ULMFiT)}
A single model architecture that is used for both LM pretraining and downstream fine-tuning was first introduced in ULMFiT \cite{ulmfit}. This allows the weights learnt during pretraining to be reused instead of constructing a new task-specific model. Howard and Ruder suggested that LM overfits to small datasets and suffers catastrophic forgetting when directly fine-tuned to a classifier. Hence, the ULMFiT approach was proposed to attempt to effectively fine-tune the AWD-LSTM \cite{awdlstm} model. ULMFiT is a 3-stage training method consisting of LM pretraining, LM fine-tuning, and classifier fine-tuning. They also proposed novel techniques such as discriminative fine-tuning, gradual unfreezing, and slanted triangular learning rates for stable fine-tuning. 
\vspace{-1.5mm}
%The model has an embedding size of 400, 3 layers of 1150 hidden activations per layer, and a BPTT batch size of 70. 

\subsection{Embeddings from Language Models (ELMo)}
Traditional monolithic word embeddings such as word2vec \cite{mikolov2013a} and GloVe \cite{pennington2014} fails to model context-dependent meanings of a word. Hence, ELMo \cite{elmo} produces contextualized word embeddings by utilizing a pretrained biLM as a fixed feature extractor and incorporate its embedding representation as features into another task-specific model for downstream tasks. The authors suggested that combining the internal states of the LSTM layers allows for rich contextualized word representations on top of the original context-independent word embeddings. 
\vspace{-1.5mm}
%The bidirectional LM constructed by combining forward and backward LMs would then have its weights frozen and the new model would then be trained as a normal classifier for classification tasks. 
%In the original paper, the bidirectional language model from which ELMo embeddings are obtained from 2 bi-LSTM layers with 4096 units and 512 dimension projections and a residual connection from the first to second layer. 

\subsection{Generative Pretrained Transformer (GPT)}
Sequential computation models used in sequence transduction problems \cite{bahdanau2014neural,cho2014,sutskever2014} forbid parallelization in the training examples. The transformer \cite{vaswani2017} is the first transduction model based solely on self-attention to draw global dependencies between input and output, eliminating the use of recurrence and convolutions. OpenAI introduced GPT \cite{openai-gpt} by extending the idea to multi-layer transformer decoder for language modeling. Additionally, LM fine-tuning and classifier fine-tuning are done simultaneously by using LM as an auxiliary objective. The authors suggested that this improves the generalization of the supervised model and accelerates convergence.
\vspace{-1.5mm}

\subsection{Bidirectional Encoder Representations from Transformers (BERT)} \label{section:bert-background}
ULMFiT \cite{ulmfit} and GPT \cite{openai-gpt} use a unidirectional forward architecture while ELMo \cite{elmo} uses a shallow concatenation of independently trained forward and backward LMs. With criticism on the standard unidirectional LMs as suboptimal by severely restricting the power of pretrained representations, BERT \cite{bert} was proposed as a multi-layer transformer encoder designed to pretrain deep bidirectional representations by jointly conditioning on both left and right context in all layers. Since the standard autoregressive LM pretraining method is not suitable for bidirectional contexts, BERT is trained on masked language modeling (MLM) and next sentence prediction (NSP) tasks. MLM masks 15\% of the input sequences at random and the task is to predict those masked tokens, requiring more pretraining steps for the model to converge. The output of the special first token is used to compute a standard softmax for classification tasks. 
\vspace{-1.5mm}

\section{Dataset} \label{section:dataset}
%In this section, the details of the datasets used in this study is discussed.

\subsection{Pretraining Dataset}
To collect our Thai social media corpus data, we extracted non-sensitive information from all threads from \emph{Pantip.com} since $1^{st}$ January 2013 up until $9^{th}$ February 2019 using our implementation of the Scrapy Framework \cite{scrapy}. A total of $8,150,965$ threads were extracted. As discussed in Section~\ref{section:data_preprocessing}, data preprocessing techniques are applied to the corpus. Length filtering and language filtering filtered down the threads to $5,524,831$ and $5,487,568$ respectively. After preprocessing, tokenization, and postprocessing the data, we divided our pretraining dataset into 3 parts: $5,087,568$ threads for training, $200,000$ threads for validation, and $200,000$ threads for testing. The train dataset, validation dataset, and test dataset has a total of $1,262,302,083$ tokens, $4,701,322$ tokens, and $4,588,245$ tokens respectively. By comparison, our pretrain dataset is more than 31 times larger than the Thai Wikipedia Dump with respect number of tokens, which is only on the order of 40M tokens for the training set. 
\vspace{-1.8mm}

\subsection{Benchmarking Dataset} \label{section:downstream}
Two Thai social text classification tasks were chosen to benchmark the models for extrinsic model evaluation as shown in Table~\ref{table:dataset}. Since both are originally Kaggle competitions, the Kaggle evaluation server will be used for benchmarking. 
\vspace{-1.8mm}
\subsubsection{Wongnai Challenge: Rating Review Prediction}
First initiated as a Kaggle competition, the Wongnai Challenge is to create a multi-class classification sentiment prediction model from textual reviews. As an emerging online platform in Thailand, Wongnai holds a large user base of over 2 million registered users with a surplus of user-written reviews accompanied by a rating score ranging from 1 to 5 stars. This is challenging due to the varying user standards, corresponding to shifting weighted importance of each sentiment in mixed reviews. 
\vspace{-1.8mm}
\subsubsection{Wisesight Sentiment Analysis}
The Wisesight Sentiment Analysis is a private Kaggle competition where the task is to perform a multi-class classification on 4 categories: positive, negative, neutral, and question. Wisesight, a social data analytics service provider, provides data from various social media sources with various topics on current internet trends. It should be noted that the topics and the source of the data are much more diverse than that of Wongnai. 

\begin{table}
\centering
\sisetup{round-mode=places,detect-weight=true,detect-inline-weight=math}
\caption{Datasets, tasks, number of classes, train and test examples, and the average example length measured in tokens. The OOV rate is measured with respect to the original vocabulary of the pretraining corpus. } \label{table:dataset} 
    \begin{threeparttable}
        \begin{tabular}{ l l c c c S[round-precision=2] c  }
        \toprule
        Dataset & Task & Classes & Train & Test & OOV & Average Length \\
        \midrule
        Wongnai   & Sentiment Classification & 5 & 40k   & 6.2k & 0.710\% & \si{126\pm124}\\
        Wisesight & Sentiment Classification & 4 & 26.7k & 3.9k & 2.685\%  & \si{27\pm44}\\
        \bottomrule
        \end{tabular}
    \end{threeparttable}
    %\vspace{-3mm}
\end{table}

\section{Experimental Setup} \label{section:setup}
\subsection{Implementation Details}

\subsubsection{ULMFiT}
We used the same model hyperparameters as the popular Thai GitHub repository \textit{thai2fit} \cite{thai2fit}: the base model is a 4-layer AWD-LSTM with $1,550$ hidden activation units per layer and an embedding size of 400. A BPTT batch size of 70 was used. We applied dropout of 0.25 to output layers, 0.1 to RNN layers, 0.2 to input embedding layers, 0.02 to embedding layers, and weight dropout of 0.15 to the RNN hidden-to-hidden matrix. %Maximum learning rate was set to 0.001 for all 3 epochs. 
\vspace{-1.5mm}
\subsubsection{ELMo}
We used the same biLM architecture from the original implementation \cite{elmo} with all default hyperparameters, where the LM is a 2-layer biLSTM with 4096 units and 512 dimension projections with another static character-based representations layer with convolutional filters. For both downstream tasks, a 3-layer biLSTM was used with 256 hidden units as the task-specific model.
\vspace{-1.5mm}
\subsubsection{GPT}
Default configurations of \cite{openai-gpt} were used. The resulting model has 12 layers of transformer each with 12 self-attention heads and 768-dimensional states. We used learnt position embeddings and a maximum sequence length of 256 tokens.
\vspace{-3.5mm}
\subsubsection{BERT}
We used the publicly available $BERT_{BASE}$ unnormalized multilingual cased model, which has a hidden size of 768, 12 self-attention heads, and 12 transformer blocks. Note that the $BERT_{BASE}$ was chosen to have identical hyperparameters as GPT for comparative purposes.  %The total number of parameters is 110M.
\vspace{-2.5mm}
% \subsection{Benckmark Tasks}
% One of the main challenges in building a language model is the method to benchmark the model due to the unsupervised nature of the task. Two main categories for language model benchmarking is the intrinsic evaluation and extrinsic evaluation. Intrinsic evaluation is to evaluate the model based on the metrics inside the model itself, which is the perplexity value--calculated through the exponential of the cross entropy loss. Minimizing the perplexity value is equivalent to maximizing the probability or model's confidence of the next word. However, this is very model-dependent and dataset-dependent, and is generally a bad approximation for the evaluation of the model's performance due to the fact assumption that the testing data is exactly the same as the training data. Another method of comparing language models is extrinsic evaluation, which is to directly compare the LMs with the downstream tasks such as: sentiment analysis, text classification, textual entailment, and question answering. This is the most unbiased way to compare LM’s, but is time-consuming and requires labeled data.

%We pretrain all LMs except BERT for 3 epochs. BERT is pretrained for 1 million steps. All models are trained with the same vocabulary of 80k tokens.

\subsection{Evaluation Metrics}
A total of 4 tasks were evaluated: the proposed data preprocessing technique in Section~\ref{section:data_preprocessing}, LM pretraining, LM fine-tuning, and classifier fine-tuning.
% Due to the unsupervised nature of language modeling, it is challenging to find a method to benchmark the language models. Hence, 
We chose to benchmark on the easiness to train each model (speed and number of epochs), the intrinsic evaluations (perplexity), and the extrinsic evaluations (downstream classification tasks). In addition, an ablation study of limited corpus data is compared to see the performance of each model in smaller data scenarios.
\vspace{-2.5mm}
\subsubsection{Data Preprocessing}
To benchmark the quality of our unique data preprocessing techniques for Thai social media corpus, we sampled a thread from each dataset and request expert Thai native speakers to help tokenize the samples. At the time of writing, there is no standard corpus for benchmarking the task of colloquial Thai word segmentation. Each character in the thread is labeled as 1 (beginning of word) or 0 (intra-word character). The precision, recall, and F1 score is calculated based on the performance of segmenting each character, where true positives are the correctly segmented beginning of word.  The default pyThaiNLP tokenizer \cite{pythainlp} Maximum Matching (newmm) is compared between with and without our data preprocessing methods. Unfortunately, labeling tokenization dataset in Thai language requires large amount of effort. Therefore, more extensive experiments will be conducted in the future. 

\vspace{-2.75mm}
\subsubsection{Language Model Pretraining}
Pretraining a language model is the most expensive process in the transfer learning workflow. This task is generally performed only once before fine-tuning on a target task. With minimal hyperparameter tuning, we evaluated the pretraining process on: (1) the speed of training in each epoch and (2) the intrinsic perplexity value. Although with the ambiguity that comes with intrinsic metrics, perplexity is one of the traditional methods in LM evaluation. It measures the confidence of the model on the observed sequence via exponentiation of the cross-entropy loss, where cross-entropy loss is defined as the negative sum of the mean LM log-likelihood. Note that this definition applies to different levels of granularity. Due to resource constraints, each model was pretrained for a fixed number of epochs. An NVIDIA P6000 is used to pretrain each model, and the appropriate batch size was selected such that it maximizes the GPU VRAM of 24 GB. The models were trained for 3 epochs and the best performing model was selected. However, since BERT trains using MLM and is able to learn just 15\% of the corpus during 1 epoch, we decided to train for the standard 1 million steps \cite{bert} (equivalent to around 6.5 epochs). 
\vspace{-2mm}
\subsubsection{Language Model Fine-tuning}
Each model was benchmarked on the number of epochs used and the total time until convergence. This process aims to learn the slight differences in data distribution of the target corpus.
% Due to the modest size of the corpus, the LM fine-tuning process can be done very quickly. 
The models overfit easily due to the modest size of the corpus, thus each LM was fine-tuned until early stopping. 
\vspace{-1.75mm}
\subsubsection{Classifier Fine-tuning}
 In this paper, we reported each downstream task performance following the metric used in each Kaggle competition. Wongnai Rating Review Challenge and Wisesight Sentiment Analysis both use classification accuracy for evaluation, which is calculated by the proportion of correctly classified samples out of all the samples. Kaggle ranks the competitors' final standings with the private score, hence this will be used as the benchmark. 
 
%  The LMs are then changed to classification models with respect to each of their unique methods. The models are the fine-tuned and fit on the downstream tasks .

\section{Results} \label{section:result}
In this section, we first report the results of our unique preprocessing methods, followed by the results of pretraining the data. We then compare the results of ULMFiT, ELMo, GPT, and BERT with the previous state-of-the-art models in the Thai NLP research community from the Kaggle competition benchmarks. 

%%%%%%%%%%%%%%%%%%%%%%%%%%%%%%%%%%%%%%%%%%%%%%%%%%%%%%%%%%%%%%%%%%%%%
\subsection{Data Preprocessing}

 Results are shown in Table~\ref{table:token_acc}, where our preprocessing method allows the default pyThaiNLP maximum matching (MM) tokenizer to more precisely segment noisy social media data. This is due to the lower false positive tokens segmented by the noisiness of the data, where most of the spams and repetitions are preprocessed correctly. With more comprehensive vocabulary, it allows the tokenizer to segment short colloquial words more accurately. Note that this does not account for the supposed increased comprehension of the models from standardizing the data.
\vspace{-2mm}
\begin{table}
\centering
\sisetup{
        round-mode=places,
        detect-weight=true,
        detect-inline-weight=math,
        table-text-alignment=right
}
\caption{Tokenization Precision, Recall, and F1-score} \label{table:token_acc}
    \begin{threeparttable}
        \begin{tabular}{ l S[round-precision=2] S[round-precision=2] S[round-precision=2] }
        \toprule
        {Tokenizer} & {Precision} & {Recall}& {F1-score} \\
        \midrule
        %\multirow{2}{*}{MM*+Our Preprocessing} & 95.83\% & \bfseries 98.65\% & \bfseries 97.22\% \\
        {MM+Our Preprocessing} & \bfseries 95.83\% & 98.65\% & \bfseries 97.22\%\\
        %\midrule
       %\multirow{2}{*}{MM* } & 0 & \bfseries 96.04\% & 97.39\% & 96.71\% \\ 
       {MM} & 96.04\% & \bfseries 97.39\% & 96.71\% \\
        \bottomrule
        \end{tabular}
    \end{threeparttable}
    %\vspace{-4mm}%Put here to reduce too much white space after your table 
\end{table}

%%%%%%%%%%%%%%%%%%%%%%%%%%%%%%%%%%%%%%%%%%%%%%%%%%%%%%%%%%%%%%%%%%%%%
\subsection{Language Model Pretraining}

%From Table~\ref{table:pretrain_time}, AWD-LSTM with ULMFiT requires the least amount of time per epoch due to least amount of parameters of only 24M. Since ELMo is the only model that is trained with 2 P6000 GPUs, time per epoch of 52 hours is not fair. Even though ELMo only has 94M parameters compared with 117M of OpenAI GPT and 110M of BERT, its training time is longer due to the fact that ELMo also makes character-level convolution.
From Table~\ref{table:pretrain_time}, AWD-LSTM with ULMFiT requires the least amount of time per epoch and the least total time, 100 hours and 33 hours respectively. Due to resource scheduling limitations, ELMo is trained with 2 P6000 GPUs, making the total time and the time per epoch much lower than the supposed value. With character-level convolutions and character-based operations, ELMo training time should be the longest amongst all the LMs. Transformer-based models require time around more than 1.5x of ULMFiT.

\begin{table}
\centering
\caption{Model Pretraining Time. $t_{epoch}$ is the time used per epoch.}\label{table:pretrain_time}
\begin{threeparttable}
\begin{tabular}{ l | c c c }
\toprule
 {Model} & $t_{epoch}$\\
 \midrule
 \textbf{ULMFiT} 
    &  \textbf{33 hr}\\  
 biLM(ELMo) (2 GPU)
    & 52 hr\\
 GPT $seq_{max}=256$ 
     & 55 hr \\
 BERT $seq_{max}=256$ 
     & 49 hr\\
 \bottomrule
\end{tabular}
\end{threeparttable}
\vspace{-2mm}
\end{table}

The training loss and perplexity are shown in Table~\ref{table:perplexity}. BERT has the lowest word-level cross-entropy loss with $15.3857$ MLM perplexity. This is expected due to the difference of the MLM prediction task with fully visible beginning and ending context, providing more contextual information to predict the masked word as compared with traditional forward and backward models. In the domain of traditional autoregressive models, GPT has a lower perplexity than ULMFiT. ELMo is not compared to other models due to prediction granularity difference and is reported as is. 

\begin{table}
\centering
\sisetup{
        round-mode=places,
        detect-weight=true,
        detect-inline-weight=math,
        table-text-alignment=center
}
\caption{Training Loss and Perplexity After Pretraining} \label{table:perplexity}
\begin{threeparttable}
\begin{tabular}{ l S[round-precision=4] S[round-precision=4] }
\toprule
 {Model} & {Loss} & {Perplexity} \\
 \midrule
 ULMFiT 
    & 3.528132 & 34.06028 \\  
 GPT $seq_{max}=256$ 
    & 3.173512419 & 23.89125325 \\
 BERT MLM $seq_{max}=256$ 
    & \bfseries 2.7334368 & \bfseries 15.38567374 \\
\midrule
 biLM(ELMo) (Character-Level) &  1.71400 & 5.55115 \\
 \bottomrule
\end{tabular}
\end{threeparttable}
\vspace{-0mm}
\end{table}

%%%%%%%%%%%%%%%%%%%%%%%%%%%%%%%%%%%%%%%%%%%%%%%%%%%%%%%%%%%%%%%%%%%%%
%% downstream
% which one [with time/epoch constraints] relatively easy and good result
\subsection{Language Model Fine-tuning}
All the language models are fine-tuned with the target corpus until they give the best result with respect to the validation loss. An NVIDIA P6000 is used for each model and the time required is presented in Table~\ref{table:finetunelm_time}. Transformer-based models are shown to overfit quicker than LSTM-based models.
\vspace{-1mm}
\begin{table}
\centering
\caption{Language Model Fine-tuning Time. $t_{total}$ is the total time used and $t_{epoch}$ is the time used per epoch. } \label{table:finetunelm_time}
\begin{threeparttable}
    \begin{tabular}{ l | c  c  c | c  c c }
    \toprule
    Model & \multicolumn{3}{c}{Wisesight} & \multicolumn{3}{c}{Wongnai}\\ 
    &\#Epoch & $t_{total}$ & $t_{epoch}$ & \#Epoch & $t_{total}$  & $t_{epoch}$\\
     \midrule
 ULMFiT 
    & 11 & \textbf{11 min} & \textbf{1 min} & 11 & 99 min & \textbf{9 min}\\  
 biLM(ELMo)
    & 5 & 25 min & 5 min & \textbf{2} & 64 min  & 32 min\\
 GPT $seq_{max}=256$ 
    & \textbf{3} & 57 min & 19 min & 3 & 90 min & 30 min\\
 BERT $seq_{max}=256$ 
    & \textbf{3} & 36 min & 12 min & \textbf{2} & \textbf{38 min} & 19 min\\
    \bottomrule
    \end{tabular}
\vspace{-0.5mm}
\end{threeparttable}

\end{table}
%classifier ulmfit ws 13 epochs 1 min wn 6 epochs 9 min
% elmo+lstm ws 20 epochs 3 min wn 20 epochs 30 min

\vspace{-0.5mm}

\subsection{Classifier Fine-tuning} \label{section:classifier_finetune}
The results of the downstream classification tasks are shown in Table~\ref{table:class_finetune2}. 
BERT with our pretraining data outperforms all existing models on the private set of Wongnai and Wisesight and obtains 0.9\% and 3.2\% respective absolute accuracy improvement over the state-of-the-art. 
Absolute accuracy improvements on all models and tasks are obtained when pretrained with our Thai Social Media data instead of the Thai Wiki Dump. 
%Comparing the architecture ULMFiT in both tasks, there are improvements of around $2\%$ after pretraining on our data. For BERT, after increasing the sequence length and pretraining on our data, improvements of around $5\%$ can be seen in the Wongnai task. Although ELMo has relatively lower results compared to the other LM architectures, this is suspected to be due to the simplistic nature of the biLSTM task-specific model. 

%The classification results are shown in Table~\ref{table:class_finetune2}, where all of our models pretrained on Thai social media data outperforms \textit{thai2fit}\cite{thai2fit} supervised model. For the baseline model, we trained a 4-layer biLSTM with 256 hidden units and randomly initialized word embedding weights. When the biLSTM is given ELMo-enhanced input, we gain additional 5\%  and 2\% on private and public score respectively. Finally, we used logistic regression to create additional training data for OpenAI GPT by predicting on the test set. By averaging predictions from logistic regression and OpenAI GPT, we improved the result by almost 1\% on the private test set.

\begin{table} [h]
    \centering
    \caption{Classifier Fine-tuning Results. Our models are compared to other models: the baseline that predicts the most frequent label, the latest Kaggle competition winner, and public github repositories. The public leaderboard and private leaderboard are calculated with approximately 30\% and 70\% of the test data respectively.} \label{table:class_finetune2}
    \sisetup{
        round-mode=places,
        detect-weight=true,
        detect-inline-weight=math,
        table-format=1.4
    }
    \begin{threeparttable}
        \begin{tabular}{ l |@{\hskip 0.3cm} S[round-precision=4]@{\hskip 0.3cm} S[round-precision=4]@{\hskip 0.3cm} |@{\hskip 0.3cm} S[round-precision=4]@{\hskip 0.3cm}  S[round-precision=4]}
            \toprule
            {Model} & \multicolumn{2}{c}{Wisesight (Acc.)} & \multicolumn{2}{c}{Wongnai (Acc.)}\\
            & {Private} & {Public} & {Private} & {Public} \\
            \midrule
            Baseline 
                & 0.58089 & 0.60439
                & 0.4785 &  0.4785\\
            Kaggle Best 
                & 0.7597 & 0.7532
                & 0.59139 & 0.58139\\
            fastText \cite{thai2fit}
                & 0.6131 & 0.6314
                & 0.5145 & 0.5109\\
            LinearSVC \cite{thai2fit}
                & {-} & {-}
                & 0.5022 & 0.4976\\
            Logistic Regression \cite{thai2fit}
                & 0.7499 & 0.7278
                & {-} & {-}\\
            \midrule
            \multicolumn{5}{l}{\textbf{\textit{Thai Wiki Dump Pretraining}}} \\
            ULMFiT \cite{thai2fit}
                & 0.74194 & 0.71259
                & 0.59313 & 0.60322 \\
            ULMFiT Semi-supervised \cite{thai2fit}
                & 0.75968 & 0.73372                
                & {-} & {-} \\
            BERT $seq_{max}=128$ \cite{thaikeras}
                & {-} & {-}                
                & 0.56612 & 0.57057\\
            \midrule
            \multicolumn{5}{l}{\textbf{\textit{Ours (Thai Social Media Pretraining)}}} \\
            ULMFiT 
                & 0.75859 & 0.73457               
                & 0.62030 & \bfseries 0.64086\\
            biLSTM
                & 0.63662 & 0.62130
                & 0.47731 & 0.49462 \\
            ELMo+biLSTM
                & 0.68657 & 0.64497                
                & 0.53096 & 0.52258\\
            GPT $seq_{max}=256$
                & 0.76692 & \bfseries 0.75401
                & 0.60879 & 0.61451 \\
            %GPT $seq_{max}=256$ Semi-supervised
            %    & {-} & {-}
            %    & \bfseries 0.77452 & 0.74725\\
            BERT $seq_{max}=256$
                & \bfseries 0.76909 & 0.74387
                & \bfseries 0.62514 & 0.62311\\
            \bottomrule
        \end{tabular}
    \end{threeparttable}
    \vspace{-1.5mm}
\end{table}

\subsection{Limited Pretraining Corpus}
% compare with thai2fit and see ~almost the same
%which one perfirm well on limited size of data
We also investigated the performance of the models in the scenario where the pretraining corpus is limited. This result reflects the learning ability of the models in a language where training data is scarce. We randomly sampled a total of 40M tokens (equivalent to around 234K threads) from the dataset used in our previous experiments. ULMFiT, ELMo, and GPT are trained for 3 epochs while BERT is trained for 30k steps (equivalent to approximately 6.5 epochs on this data). Table~\ref{table:limited_corpus} shows that ULMFiT and GPT perform considerably well. On the other hand, adding ELMo to LSTM input shows little improvement. This means that ELMo requires a larger corpus to be effective. Although BERT performs well on the Wisesight dataset, it has a drop in performance on Wongnai dataset.
\vspace{-2mm}
\begin{table} [h!]
    \centering
    \caption{Limited Pretraining Corpus Results. The public and private scores are calculated with approximately 30\% and 70\% of the test data respectively.} \label{table:limited_corpus}
    \sisetup{
        round-mode=places,
        detect-weight=true,
        detect-inline-weight=math,
        table-text-alignment=right,
        table-format=2.5
    }
    \begin{threeparttable}
        \begin{tabular}{ l | S[round-precision=4] S[round-precision=4] | S[round-precision=4]  S[round-precision=4]}
            \toprule
            {Model} & \multicolumn{2}{c}{Wisesight (Acc.)} & \multicolumn{2}{c}{Wongnai (Acc.)}\\
            & {Private} & {Public} & {Private} & {Public} \\

            \midrule
         ULMFiT 
            & 0.73579 & 0.71428 & 0.59843 & \bfseries 0.62903 \\  
        biLSTM
            & 0.63662 & 0.62130 & 0.47731 & 0.49462 \\
         biLSTM + ELMo 
            & 0.64893 & 0.60946 & 0.48791 & 0.47526\\
         GPT $seq_{max}=256$ 
            & 0.69308 & 0.70752 & \bfseries 0.61109 & 0.61021\\
         \bfseries BERT $seq_{max}=256$ 
            &  \bfseries 0.74665 &  \bfseries 0.72442 & 0.56504 & 0.55161\\
            \bottomrule
        \end{tabular}
    \end{threeparttable}
\end{table}
\vspace{-1mm}

\section{Conclusion} \label{section:conclusion}
% talk about results 
Our work shows that by using our unique data preprocessing methods and our pretraining social media data, we can improve the performance of the LMs in the downstream tasks. The improvement of all models from pretraining data of the same domain suggests that pretraining data has a significant impact on LM performance. Moreover, the possibility for LM pretraining on a noisy corpus shows the ability of the models to learn in spite of the quality of the data. 
% With the exponential growth of online data, we expect to see more large pretraining corpus based on user-generated content and generating high-performance models resistant to noise.

Results-wise, BERT is the best performing model with respect to classification accuracy. It can achieve state-of-the-art results on both of the benchmarking downstream tasks. However, it has unstable performance on downstream tasks when pretrained on a small corpus and uses a lot of pretraining time. If speed and ease of training are the main considerations, we recommend using AWD-LSTM with ULMFiT due to its speed of pretraining and fine-tuning, while the results are still on par with transformer-based models. Although OpenAI GPT shows promising results with acceptable pretraining speed, it is overshadowed by other models in both aspects. Finally, although ELMo shows significant improvements when compared with the baseline biLSTM, it places a dependency on designing a powerful task-specific model to achieve good performance.

\vspace{-1mm}
\section*{Acknowledgements} \label{section:ack}
In the making of the paper, the authors would like to acknowledge Mr. Can Udomcharoenchaikit for his continuous and insightful research suggestions until the completion of this paper. 
\vspace{-1mm}
\bibliographystyle{splncs04}
\bibliography{references}

\end{document}